%% file: Main.tex
\documentclass{IET-Conf-Paper}

\usepackage{multirow}
\usepackage[hyphens]{url}
\usepackage{natbib}  % or other citation packages
\begin{document}

\title{Optimal Scheduling of Electric Vehicle Charging with Deep Reinforcement Learning Considering End Users Flexibility}

\author{Christoforos Menos-Aikateriniadis\ad{1,}\ad{2}\corr, Stavros Sykiotis\ad{3}, Pavlos S. Georgilakis\ad{1}}

\address{\add{1}{School of Electrical and Computer Engineering, National Technical University of Athens, Athens, Greece}
\add{2}{Intracom S.A. Telecom Solutions, Telco Software Dpt, 19.7 km Markopoulou Ave., Peania, Athens, Greece}
\add{3}{School of Rural, Surveying and Geoinformatics Engineering, National Technical University of Athens, Athens, Greece}
\email{christoforosmenos@mail.ntua.gr}}


\begin{abstract}
The rapid growth of decentralized energy resources and especially Electric Vehicles (EV), that are expected to increase sharply over the next decade, will put further stress on existing power distribution networks, increasing the need for higher system reliability and flexibility. In an attempt to avoid unnecessary network investments and to increase the controllability over distribution networks, network operators develop demand response (DR) programs that incentivize end users to shift their consumption in return for financial or other benefits. Artificial intelligence (AI) methods are in the research forefront for residential load scheduling applications, mainly due to their high accuracy, high computational speed and lower dependence on the physical characteristics of the models under development. The aim of this work is to identify households' EV cost-reducing charging policy under a Time-of-Use tariff scheme, with the use of Deep Reinforcement Learning, and more specifically Deep Q-Networks (DQN). A novel end users flexibility potential reward is inferred from historical data analysis, where households with solar power generation have been used to train and test the designed algorithm. The suggested DQN EV charging policy can lead to more than 20\% of savings in end users electricity bills.
\end{abstract}

\maketitle

\section{Introduction}\label{sec:intro}

The main focus of an electric power system is to ensure security of supply with the least possible cost \cite{Karafotis2020Evaluation}. On a residential level, the penetration of decentralized energy resources, such as photovoltaic systems (PV), local energy storage and electric vehicles (EV), have increased the difficulty in load and generation forecasting, and have consequently created the need for higher flexibility on both the demand and the supply side. In addition, this necessity is expected to grow more in the next decade, in line with the ongoing penetration of EV on a residential level. Based on the latest IEA's Global EV Outlook, EV global electricity demand can account for 1,100 TWh, around 4\% of total global demand by 2030 \cite{IEA_EV}. Demand response (DR) can offer such flexibility in residential, low voltage networks by controlling large flexible household loads, such as EV, through the end users’ home energy management systems (HEMS). Incentivizing end users to shift their consumption in different hours can be beneficial for network operators to tackle voltage, frequency and demand-related issues. In the meantime, consumers can decrease their electricity bills or be rewarded in other ways (incentives) by the network operators when agreeing to participate on a DR event. 

Smart meters are a key component of such grid monitoring systems, since real data can be used to train AI models that can accurately analyze aggregated consumption signals of domestic appliances \cite{Sykiotis2022electricity}, infer battery's State of Charge (SoC) \cite{OverviewSoC}, identify consumption patterns and optimize the scheduling of residential resources in demand response applications \cite{Rajasekhar2020Survey,Mabina2021Sustainability}. Reinforcement Learning (RL) and nature-inspired algorithms, such as genetic algorithm (GA) and particle swarm optimization (PSO), are commonly used for optimally scheduling residential energy resources given the lower modelling complexity and the lack of linearity in the problem formulation \cite{Antonopoulos2020Artificial, Menos2022Particle}. However, determining optimal solutions with model-based approaches can become challenging when stochastic variables, such as end users consumption behavior, are considered \cite{Liu2020Optimization}. In this case, system model information is required, modeling complexity increases and there is a lack of scalability and adaptability in the designed algorithmic solutions. On the contrary, Reinforcement Learning shows a more dynamic character than nature-inspired methods, since RL can be updated while in operation and continuously learn from past experiences. 

Many authors have investigated and evaluated the performance of Reinforcement Learning in residential DR applications, since model-free RL methods can be used without the need of explicit mathematical formulation to model end users' consumption habits \cite{Vazquez2019Reinforcement}. More specifically, the use of Deep Reinforcement Learning for EV scheduling has attracted increasing interest in the last few years. Both works \cite{Li2020SafeDQN, Wan2019EV_DRL} use a Deep Q-learning (DQN) algorithm to optimize EV charging scheduling for residential DR, considering past electricity prices. Even if both works show high performance on electricity costs reduction, end users commuting habits and therefore their charging flexibility is modeled as a normal distribution with randomly selected variables. Similarly, in work \cite{Li2020DR}, which aims at scheduling multiple energy resources, including EV, residential consumption patterns are based on Gaussian probability density functions with consumer preferences not inferred from historical data. In work \cite{Chis2017PEV} end user driving preferences have been considered known in an attempt to identify a cost-reducing, long-term charging policy for a plug-in EV with batch, fitted Q-Iteration algorithm. On the contrary, in work \cite{Shuvo2022HERS} electricity costs and user discomfort have been jointly minimized with a DQN algorithm, where end user feedback is included in the rewards and sensor-based human activity in the RL states. Despite the clear contributions of this work, authors did not consider the contribution of renewable power generation and therefore consumption's carbon footprint in the energy model. The latter has been introduced in work \cite{Ren2022Novel} where DQN, double DQN and dueling double DQN are compared for a household energy management system that controls Heating, Ventilation, and Air Conditioning (HVAC), PV, EV and energy storage. EV charging has been based on end users' daily satisfaction, as obtained from historical data.  

The above literature review indicates that there has been an increasing interest in EV scheduling optimization for DR applications with the use of Deep RL. However, in the majority of existing works, end users charging availability and flexibility are being modeled as stochastic variables without any correlation with historical consumption data. Furthermore, the contribution of solar power generation on EV optimal charging scheduling has been rarely witnessed in the literature reviewed. 

In this work a novel framework for EV charging cost minimization that considers end user flexibility and solar power generation, is proposed with the use of DQN. The contributions of this paper can be summarized as follows:
\begin{itemize}
    \item Provides a thorough residential EV load scheduling model through solar power generation, expanding on top of the rather limited existing research on this area. Different households and days with solar power generation are included in the training and test data sets.
    \item Utilizes a flexibility potential reward inferred from real household measurements. Instead of modeling end users charging habits with a probability density function, as in works \cite{Li2020SafeDQN,Chis2017PEV}, a charging availability index is integrated in the Deep RL reward function, reflecting the probability for a user to charge their EV at each time interval.  
    \item Formulates a constrained EV load scheduling problem, considering battery's technical characteristics and end user driving daily patterns. Both aspects are integrated into the Deep RL environment as EV Battery SoC monitoring and daily EV consumption rewards, respectively. 
\end{itemize}

The structure of this paper is as follows. Section \ref{sec:methodology} presents the methodology followed to formulate the problem from an energy and RL perspective. Section \ref{sec:results}  presents the experimental setup and compares the proposed DQN method with metered data. Section \ref{sec:conclusions} concludes the paper.

\input{sections/fig_flexindex}

\section{Methodology} \label{sec:methodology}

This work focuses on optimally charging the electric vehicle of residential EV owners. The energy system under examination consists of a household connected to the main grid, a battery electric vehicle (BEV), an EV slow charger, rooftop solar PV panels and a 'residual load' consisting of the cumulative consumption from all other appliances. Real-time optimal EV charging is formulated as a 15-minute (discrete) time step optimization problem, aiming at distributing the charging load to optimal time intervals throughout the day. At each time step $\textit{t}$ of the 24-hour modeling horizon ($\textit{T}$ = 96), the aim is to define whether it is beneficial for the end user to charge their EV as well as estimate the amount of energy required, based on electricity retail tariffs and user daily driving habits, without jeopardizing their convenience. 

User convenience can be described through a charging availability index that takes into consideration the historical daily EV charging pattern. The charging availability index expresses the frequency for an EV user to charge their vehicle at each 15-minute interval, as inferred by analysis performed on smart metered data from households in Austin, Texas, US found in the open-source data set of Pecan Street \cite{Pecan}. Fig. \ref{fig:flexindex} shows the charging availability profile for an indicative household of the dataset. The high flexibility index from 17.00 to 21.00 indicates the user's preference to charge their BEV in the respective time slots. The quantiles of the charging probability density function are computed and specific thresholds are being selected to represent low, medium and high preference periods for EV charging.

In a Reinforcement Learning context, mechanisms of the optimization task need to be defined as an environment, which produces observations, and rewards an optimizer (agent), in the form of a Deep Neural Network, depending on its actions. Optimal EV charging task is formulated as a Markov Decision Process (MDP) defined by the tuple $(\mathcal{S},\mathcal{A},R,P)$. $\mathcal{S}$ signifies the state space, i.e. the observations that the agent will evaluate to choose an action.

\input{sections/fig_overview.tex}
\noindent
The actions are obtained in the set $\mathcal{A} = \{1,0\}$, meaning that the agent can, at any given time step \textit{t}, choose either to charge the EV or remain idle. The environment will then evaluate the action of the agent, depending on specific criteria and assign a respective reward. This process is iterated until the agent learns to choose the actions that maximize its rewards by learning the optimal state-action pairs $(s,a)$ at any timepoint \textit{t}. The agent then receives the next state \textit{s'} and the same routine is repeated through the training phase. A high-level overview of this approach is illustrated in Fig. \ref{fig:high_level}.

\subsection{State}
At each time step \textit{t}, the environment produces a vector $s_{t} = (\textit{p}_{t}, \textit{P}_{t}^{PV}, \textit{P}_{t}^{non-EV}, {P}_{t,run}^{EV}, \textit{SoC}_{t}, t) $ that is passed to the optimization agent. The information included in the state space is the following \cite{synergy22}: (1) $p_{t}$ represents the Time-of-Use (ToU) electricity tariff at time $\textit{t}$, based on the load distribution (On-Peak, Mid-Peak, Off-Peak); (2) $P_{t}^{PV}$ is the power generated by the rooftop solar PV; (3) $P_{t}^{non-EV}$ denotes the total residual (non-EV) load; (4) ${P}_{t,run}^{EV}$ shows the running EV power consumption, starting from the initialization of the episode (t=1) until the current time step \textit{t}; (5) $SoC_{t}$ indicates the State of Charge at time step $\textit{t}$, which (6) holds information about the current time step of the episode.

In this work, the data are split on a daily basis to formulate episodes. Before the episode starts, the environment calculates the amount of power that the end user consumed to charge their EV ($P_{day}^{EV}$) using historical consumption data. Given that the number of EV battery charging cycles per day is unknown, it is assumed that the historically consumed daily power corresponds to a full charging cycle \cite{synergy22} and starting SoC, $SOC_{start}$, is calculated as in Equation (\ref{eq:EV_SoC}):

\begin{equation}\label{eq:EV_SoC}
      SoC_{start} =1 - \eta \frac{ P_{day}^{EV}}{4 E_{batt}}
\end{equation}
\noindent
where $\eta$ is the charging efficiency of the battery pack and $E_{batt}$ is the rated battery capacity in kWh. Equation (\ref{eq:constr_Eday}) ensures that optimized EV daily power consumption will remain similar ($\pm 5\%$) to the historical consumption $P_{day}^{EV}$ after any load shifting actions on that specific episode (day). 

\begin{equation}\label{eq:constr_Eday}
   P^{EV}_{day} \in [ 0.95\sum_{t=1}^{T} P^{EV}_t, 1.05\sum_{t=1}^{T} P^{EV}_t  ]
\end{equation}

\subsection{Action}
At every state $s_{t}$ the agent should select an action as follows:

 \begin{equation}\label{eq:constr_onoff}
   \alpha_t^{EV} = \{1, 0\}, \forall \alpha \in A, \forall t \in T
 \end{equation}
\noindent
where $\alpha_t^{EV}$ equals to 1 when the optimization algorithm suggests the BEV to charge and 0 when it is preferable to remain idle and not charge. 

In addition, RL actions are constrained by the technical characteristics and the physical properties of the EV battery pack, which are described as follows:

\begin{equation}\label{eq:EV_SoC_limits}
  {SoC_{start}} \leq SoC_{t} \leq {SoC_{max}}, \forall t \in T
\end{equation}

\begin{equation}\label{eq:EV_updateSoC}
  SoC_{t+1} = SoC_{t} +  \frac{\eta  P_{t}^{EV}}{4  E_{batt}}, \forall t \in T
\end{equation}

\begin{equation}\label{eq:EV_charging_levels}
    P_{t}^{EV} = \begin{cases}
          3.3 kW, & SoC_{min} \leq SoC_{t} \leq 0.9 \\
          1.5 kW, & SoC_{t} > 0.9 \\
         \end{cases}
         , \forall t \in T
\end{equation}
\noindent

where $SoC_{start}$, $SoC_{max}$ are the starting and maximum BEV State of Charge, $P_{t}^{EV}$ expresses the charging power consumption at a given time step $\textit{t}$ and $SoC_{t}$ expresses the State of Charge at time step $\textit{t}$, based on the charging activity as described in Equation (\ref{eq:EV_charging_levels}).

\subsection{Rewards}

Each new state $s_{t+1}$ depends on the current state $s_{t}$, on the selected action $\alpha_t^{EV}$, as well as on the reward \textit{r} associated with this action. Through the assignment of rewards, the agent can evaluate the quality of each action and learn by that. In this work, the aim is to decide which is the most cost-effective EV charging strategy considering end user flexibility potential, without violating battery's technical constraints (SoC) and user's daily driving habits (power consumption). 

To translate the aforementioned modeling process into rewards, the sub-rewards of daily power consumption ($r_{1}$), flexibility potential ($r_{2}$), cost minimization ($r_{3}$) and EV Battery SoC monitoring  ($r_{4}$) have been formulated as follows: 
\begin{equation}\label{eq:r1}
    r_1 = \begin{cases}
          1,  &  \alpha_t^{EV} = 1 \hspace{2mm} \& \hspace{2mm} P^{EV}_{t,run} \leq 1.05  P^{EV}_{day} \\
         -10, &  \alpha_t^{EV} = 1 \hspace{2mm} \& \hspace{2mm}  P^{EV}_{t,run} \geq 1.05  P^{EV}_{day} \\
        -0.5, &  \alpha_t^{EV} = 0 \hspace{2mm} \& \hspace{2mm}  P^{EV}_{t,run}  \leq 1.05 P^{EV}_{day} \\
          1,  &  \alpha_t^{EV} = 0 \hspace{2mm} \& \hspace{2mm}  P^{EV}_{t,run} \geq 1.05  P^{EV}_{day} \\
         \end{cases}
\end{equation}

 \input{sections/fig_costQx}
% \noindent

\begin{equation}\label{eq:r2}
    r_2 = \begin{cases}
         -2,  &  \alpha_t^{EV} = 1 \hspace{2mm} \& \hspace{2mm}  U^{flex}_{t}  \leq {Q}_{f}(0.25) \\
         -1, &  \alpha_t^{EV} = 1  \hspace{2mm} \& \hspace{2mm} U^{flex}_{t} \leq {Q}_{f}(0.50) \\
            0, &  \alpha_t^{EV} = 0 \\
          1, &  \alpha_t^{EV} = 1 \hspace{2mm} \& \hspace{2mm}  U^{flex}_{t} \leq {Q}_{f}(0.75) \\
          2, &  \alpha_t^{EV} = 1 \hspace{2mm} \& \hspace{2mm}  U^{flex}_{t} > {Q}_{f}(0.75) \\ 
        \end{cases}
\end{equation}
\noindent
where $U^{flex}_{t}$ represents the user charging flexibility potential, which expresses the historical consistency of the user to charge their EV on a time step \textit{t}. $Q_{f}(X)$ is the BEV charging flexibility quantile, as obtained from historical data analysis. $U^{flex}_{t}$ and quantiles $Q_{f}(X)$ are illustrated in Fig. \ref{fig:flexindex}. As can be seen in Equation (\ref{eq:r2}), the more likely it is for the end user to charge their EV on a time step \textit{t}, the highest the reward of such action should be provided by the agent. The cost minimization sub-reward ($r_3$) is given by:

\begin{equation}\label{eq:r3}
r_3 = \begin{cases}
    2,  &  \alpha_t^{EV} = 1 \hspace{2mm} \& \hspace{2mm}  C_{t} \leq {Q}_{c}(0.25) \\
    1,  &  \alpha_t^{EV} = 1 \hspace{2mm} \& \hspace{2mm}  C_{t} \leq {Q}_{c}(0.50) \\
    0,  &  \alpha_t^{EV} = 0 \\
    -1, & \alpha_t^{EV} = 1 \hspace{2mm} \& \hspace{2mm}  C_{t} \leq {Q}_{c}(0.75) \\
    -2, & \alpha_t^{EV} = 1 \hspace{2mm} \& \hspace{2mm}  C_{t} > {Q}_{c}(0.75) \\ 
    \end{cases}
\end{equation}
\noindent
where the electricity consumption cost $C_{t}$ on a time step \textit{t} is defined as:
 \begin{equation}\label{eq:obj}
   C_{t} = r_t \cdot (\alpha_t^{EV} \cdot P^{EV}_t + P^{non-EV}_t - P^{PV}_t)
 \end{equation}
\noindent
The electricity cost quantiles $Q_{c}(X)$ and the daily average electricity cost for household \#4373, as obtained from the Pecan Street data set \cite{Pecan}, are shown in Fig. \ref{fig:costQx}. $Q_{c}(X)$ has been calculated excluding all negative cost periods, to avoid setting extremely low reward thresholds. From Equation (\ref{eq:r3}) it can be seen that the greater the cost on a time step \textit{t} the worse the assigned reward will be by the agent.

Last but not least, EV Battery SoC monitoring sub-reward $r_{4}$ negatively penalizes any charging action leading to a SoC above 100 \%.

\begin{equation}\label{eq:r4}
    r_4 = \begin{cases}
          -10, & \alpha_t^{EV} = 1 \hspace{2mm} \& \hspace{2mm} SoC_{t} \geq 1 \\
          0, & otherwise
        \end{cases}
\end{equation}

After each sub-reward is being calculated, the cumulative total reward $(\mathcal{R})$ is computed as follows:

\begin{equation}\label{eq:R_total}
    R = \delta_1 \cdot r_1 + \delta_2 \cdot r_2 + \delta_3 \cdot r_3 + \delta_4 \cdot r_4
\end{equation}
\noindent
where $\delta_{k}$ is the weight factor of each sub-reward \textit{k}. In this work, it is assumed that all proposed rewards weigh the same, i.e. $\delta_{k}=0.25 \hspace{2mm} \forall k \in \{1,..,4\}$.

\section{Results}\label{sec:results}

\subsection{Experimental Setup}

Thorough data analysis on real measurements provided by the Pecan Street open source data set \cite{Pecan} has been conducted for houses in Austin, Texas. End users' historical EV charging habits have been analyzed, leading to the design of a 'flexibility potential' reward, as shown in Section \ref{sec:methodology}. Actual 2018 residential ToU rates have been considered as provided by the City of Austin \cite{AustinPrices}. As it can be seen in Table \ref{tab:tab_prices}, the ToU tariffs are categorized on Off-Peak hours, with low electricity prices at night and early morning, On-Peak hours which correspond to afternoon and evening hours and Mid-Peak hours for the remaining part of the day. It is assumed that a 24 kWh Nissan Leaf and a 90.5 \% efficient Level 2 (AC) slow charger have been used \cite{NissanLeaf}. To evaluate the proposed approach, data from house \#4373 of the Pecan dataset have been chosen, since it contains a significant amount of EV charging cycles and solar power generation \cite{Pecan}.

\input{sections/tab_prices.tex}

\subsection{Results comparison for optimal EV load scheduling}

In this work, 1,000 epochs have been used to properly train the suggested DQN model. The trained algorithm is then evaluated through a test set including days that have not been included in the training set. The comparison between the uncontrolled EV consumption and the proposed optimal EV charging policy can be seen in Fig. \ref{fig:res_weekday} and Fig. \ref{fig:res_weekend} for selected weekday and weekend test days, respectively. 

\input{sections/fig_weekday}
\input{sections/fig_weekend}

In the weekday test case, it is observed that the DQN EV proposed algorithm shifts the majority of electricity power consumption in two low-price periods: one in the morning (07:00 - 09:00) and one in the afternoon (13:00-15:00). In both periods electricity non-EV cost is rather low and flexibility potential index is non-negligible. The metered EV daily power consumption was 39.79 kW while in the DQN EV optimal strategy, power consumption is 40.2 kW, which is within the specified daily power consumption limits ($\pm 5\%$). Similarly, the DQN EV daily power consumption for the weekend test case accounts for 76.5 kW when the metered consumption was 76.1 kW. As shown in Fig. \ref{fig:res_weekend}, EV charging is being shifted from high-price time periods (e.g. 19:00 - 21:30) to lower price periods not only with high penetration of solar power generation but also with a high flexibility potential, as in periods 14:00 - 15:30 and 11:00 - 12:30. The electricity cost savings for the metered and the proposed DQN EV solutions over the test days are shown in Table \ref{tab:tab_cost}. It can be seen that the proposed, controlled solution leads to around 9.8 \% of average daily savings when compared to uncontrolled EV power consumption. These savings can reach up to 21 \% (day 09/09/2018) depending on the test case.

\input{sections/tab_cost.tex}

\input{sections/fig_weekday_bar}
\input{sections/fig_weekend_bar}

In addition, the designed DQN EV algorithm proposes an alternative EV charging pattern that, depending on the amount of generated solar power, shifts EV electrical load from high-peak periods to lower peak periods, thus reducing power peak consumption, as illustrated in Fig. \ref{fig:res_weekday_bar} and Fig. \ref{fig:res_weekend_bar} for the same weekday and weekend test cases, respectively. Although peak power consumption reduction has not been set as an optimization objective in this work, it can be observed that shifting EV load from On-Peak price periods (14:00-22:00) to different time zones within the day can reduce peak power consumption and therefore, reduce the stress on the side of the power distribution network.

\section{Conclusions} \label{sec:conclusions}

In this paper, intraday EV load scheduling optimization has been formulated as a MDP problem, when considering end users flexibility potential and solar PV power generation. A DQN EV load optimization model has been proposed, using real measurements from households in the area of Austin, Texas. A novel set of rewards has been formulated to account for end user flexibility potential, as inferred from historical user consumption data, while respecting EV battery’s technical constraints (SoC) and user’s daily habits (daily power consumption). Experimental results show that the suggested DQN EV charging solution can lead to above 20\% of electricity costs savings while reducing the stress on power distribution networks during Peak hours.   

% @misc{IEA_EV,
%   author = "IEA", 
%   title = {Global EV Outlook 2022},
%   howpublished = "\url{https://www.iea.org/reports/global-ev-outlook-2022}",
%   year = "2022",
%   note = "[Online; accessed 19-July-2008]",
%   pages = {}
% }
\section{Acknowledgements}
This project has received funding from the European Union’s Horizon 2020 research and innovation programme under the Marie Skłodowska-Curie grant agreement No 955422.

\section*{References}

\bibliographystyle{iet}
\bibliography{bibliography.bib}

\end{document}

%% file: sections/fig_flexindex.tex
\begin{figure}[ht]
    \centering
    \includegraphics[width=\columnwidth]
    {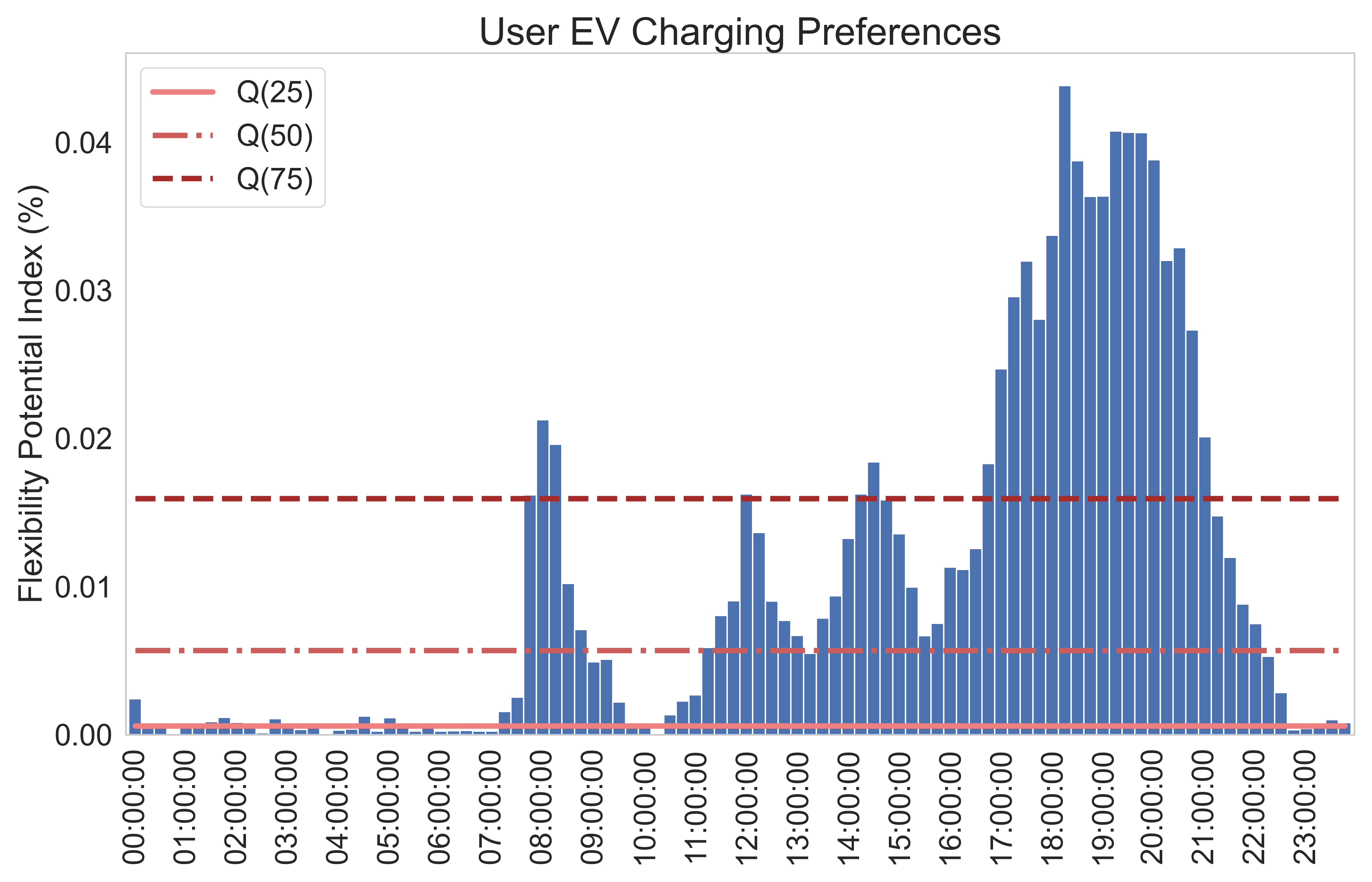}
    \caption{Charging availability index for household \#4373 in Austin, Texas. This index shows the historical EV charging frequency of user \#4373 throughout the day (\%).}
    \label{fig:flexindex}
\end{figure}

%% file: sections/fig_overview.tex
\begin{figure}[ht]
    \centering
    \includegraphics[width = \columnwidth]{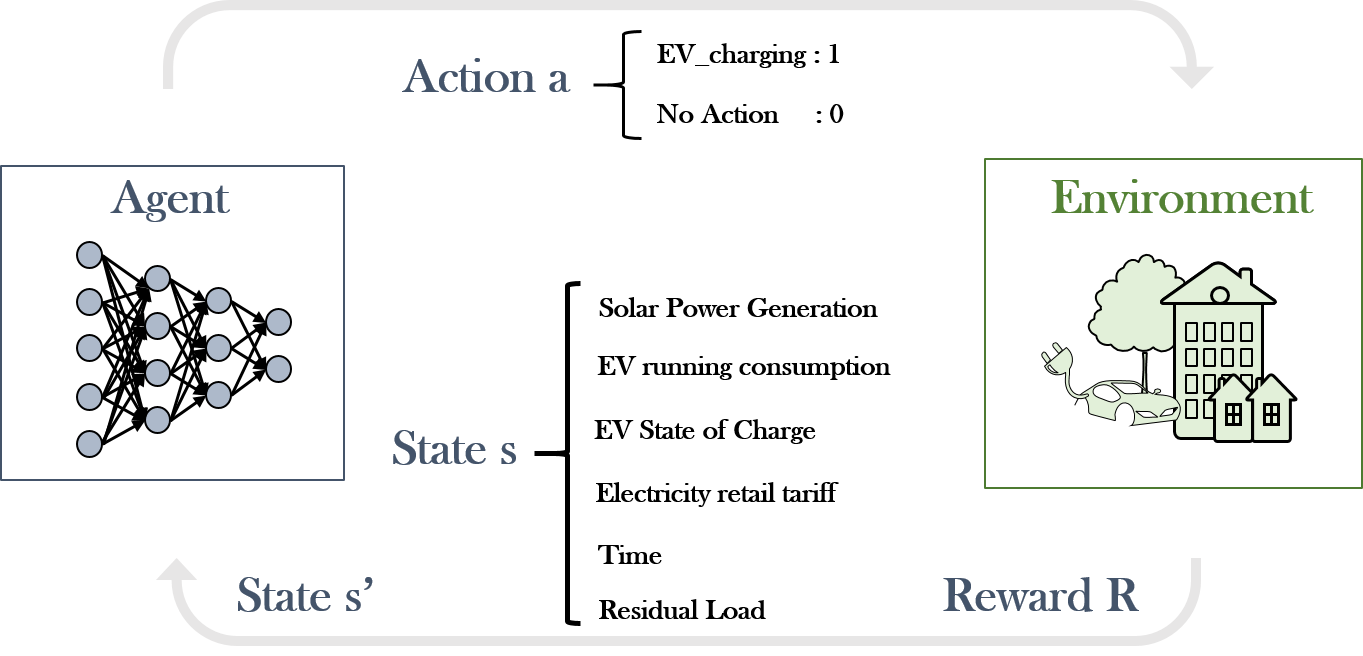}
    \caption{Overview of the proposed DQN framework for optimal EV charging. A neural network is trained by the Agent to choose an action based on a state of the environment. A reward characterizes the optimality of the action.}
    \label{fig:high_level}
\end{figure}

%% file: sections/fig_costQx.tex
\begin{figure}[ht]
    \centering
    \includegraphics[width=\columnwidth]
    {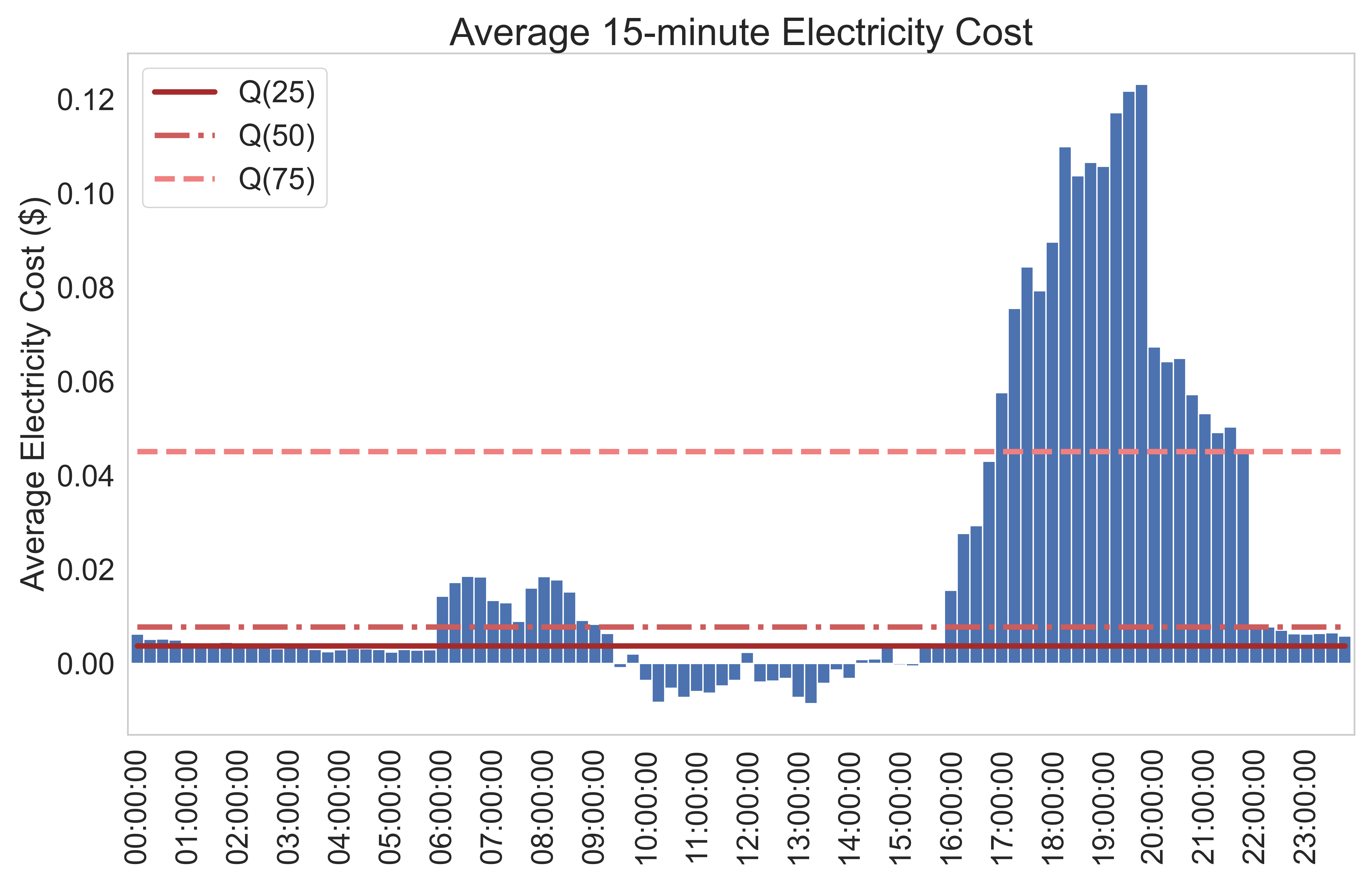}
    \caption{Average 15-minute electricity cost for household \#4373 in Austin, Texas. Negative cost 15-minute time steps are excluded from the cost quantiles calculation}
    \label{fig:costQx}
\end{figure}

%% file: sections/tab_prices.tex
\begin{table}[!ht]
\centering
\caption{2018 Time-of-use electricity tariffs for weekdays\\ in summertime period, Austin, Texas}
{\begin{tabular*}{20pc}{@{\extracolsep{\fill}}lll@{}}
\midrule
ToU Period  &   Hours  &    Electricity tariff (\$/kWh)  \\ 
\hline
Off-Peak    &   00:00 - 06:00   &    0.01188  \\
            &   22:00 - 24:00   &             \\
Mid-Peak    &   06:00 - 14:00   &    0.06218  \\ 
            &   20:00 - 22:00   &             \\
On-Peak     &   14:00 - 20:00   &    0.11003  \\
\botrule
\end{tabular*}}{}
\label{tab:tab_prices}
\end{table}

%% file: sections/fig_weekday.tex
\begin{figure}[!ht]
    \centering
    \includegraphics[width = \columnwidth]{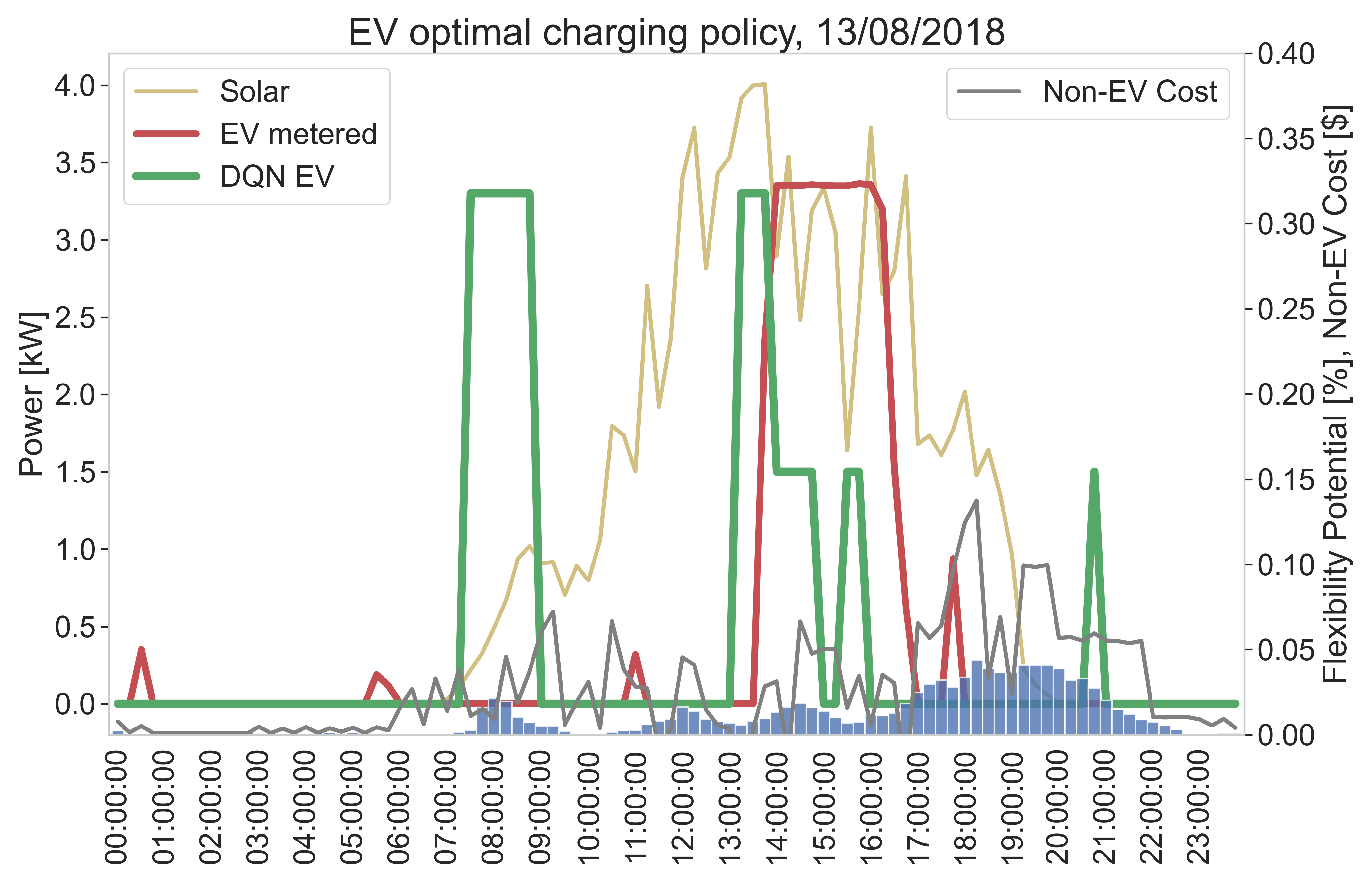}
    \caption{Weekday test case. Comparison between the DQN EV charging policy and the historically metered consumption. The proposed charging solution reduces electricity costs while respecting end-user availability.}
    \label{fig:res_weekday}
\end{figure}

%% file: sections/fig_weekend.tex
\begin{figure}[!ht]
    \centering
    \includegraphics[width = \columnwidth]{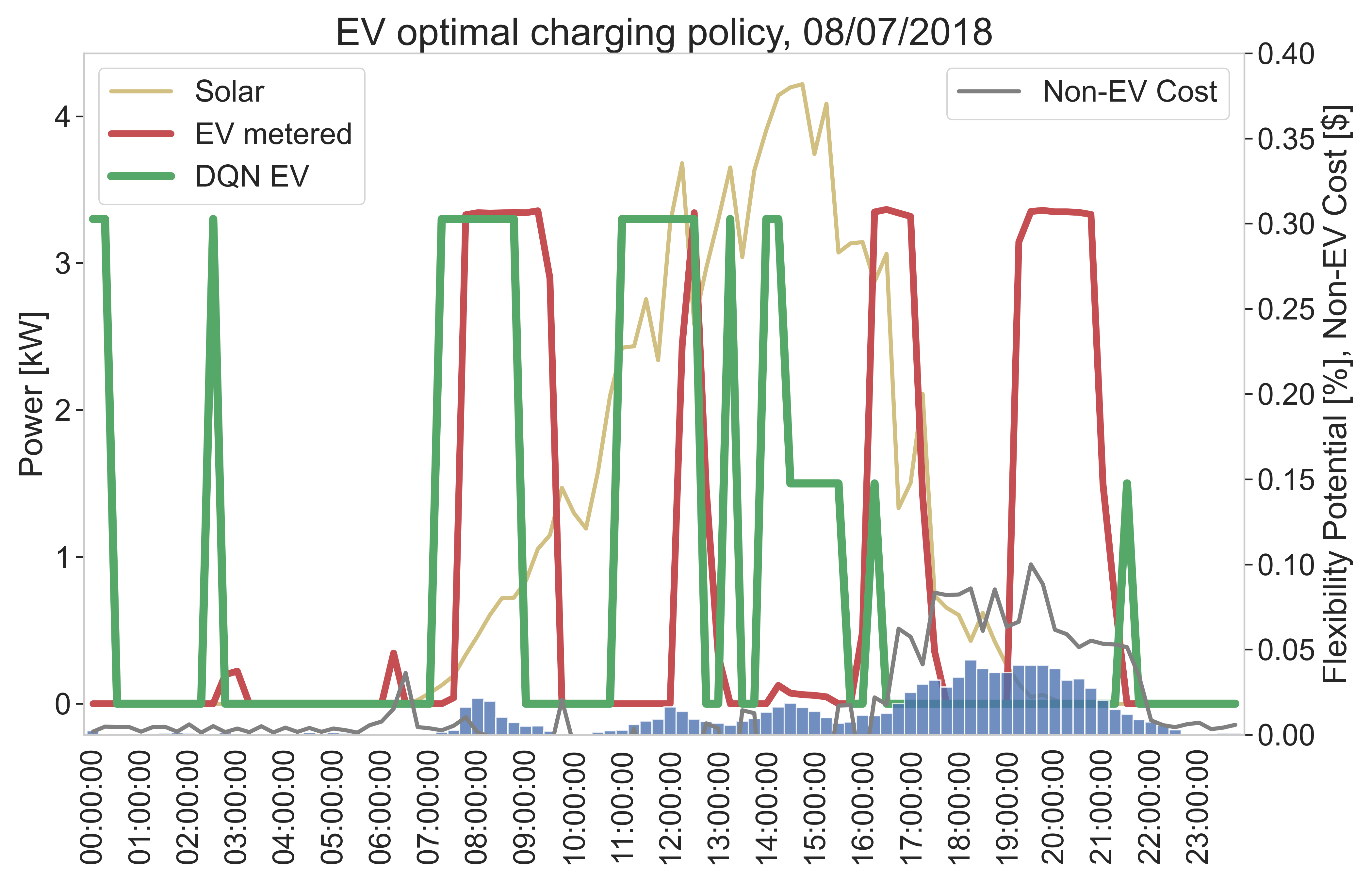}
    \caption{Weekend test case. Comparison between the DQN EV charging policy and the historically metered consumption. The proposed charging solution reduces electricity costs while respecting end-user availability.}
    \label{fig:res_weekend}
\end{figure}

%% file: sections/tab_cost.tex
\begin{table}[ht]
\centering
\caption{Daily electricity savings on selected test set days}
{\begin{tabular*}{20pc}{@{\extracolsep{\fill}}llll@{}}
\toprule
Day              &  Cost (\$)  & Cost (\$)   & Savings (\%)     \\
\midrule
27/06/2018       &    2.46     &    2.23     & 9.7               \\ 
08/07/2018       &    2.74     &    2.51     & 8.6               \\ 
12/08/2018       &    2.39     &    2.15     & 10.3              \\ 
13/08/2018       &    3.62     &    3.31     & 8.7               \\ 
19/08/2018       &    2.69     &    2.41     & 10.6              \\ 
03/09/2018       &    3.50     &    3.59     & 2.6               \\ 
08/09/2018       &    2.99     &    2.60     & 13.0              \\ 
09/09/2018       &    3.29     &    2.59     & 21.3              \\
\textbf{Average} &    2.96     &    2.67     & 9.80              \\ 
\botrule
\end{tabular*}}{}
\label{tab:tab_cost}
\end{table}

%% file: sections/fig_weekday_bar.tex
\begin{figure}[ht]
    \centering
    \includegraphics[width = \columnwidth]{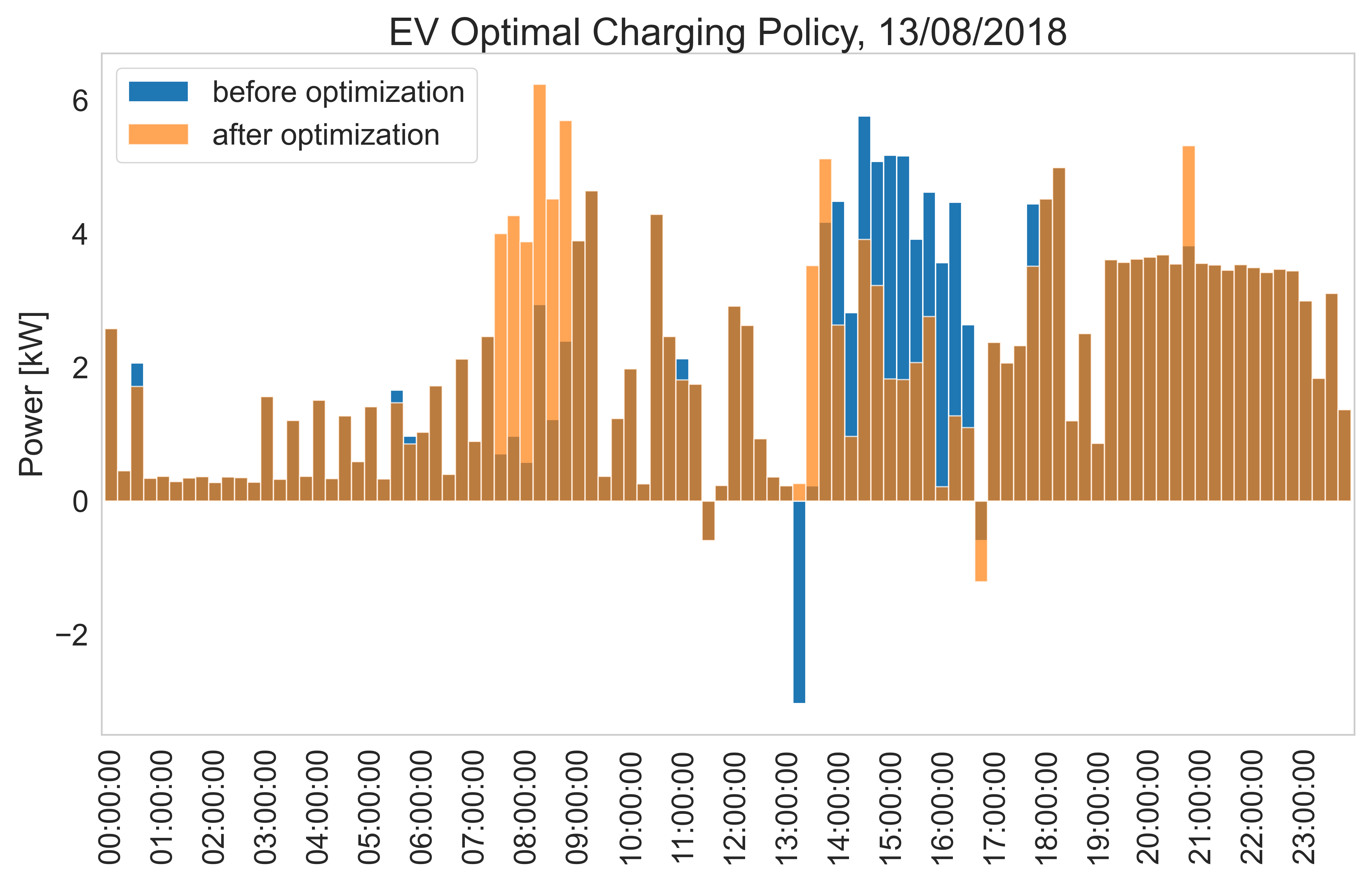}
    \caption{Weekday test case. Comparison between historical charging data and the proposed controlled EV charging policy. The system stress is being reduced due to lower power consumption on On-Peak hours}
    \label{fig:res_weekday_bar}
\end{figure}

%% file: sections/fig_weekend_bar.tex
\begin{figure}[ht]
    \centering
    \includegraphics[width = \columnwidth]{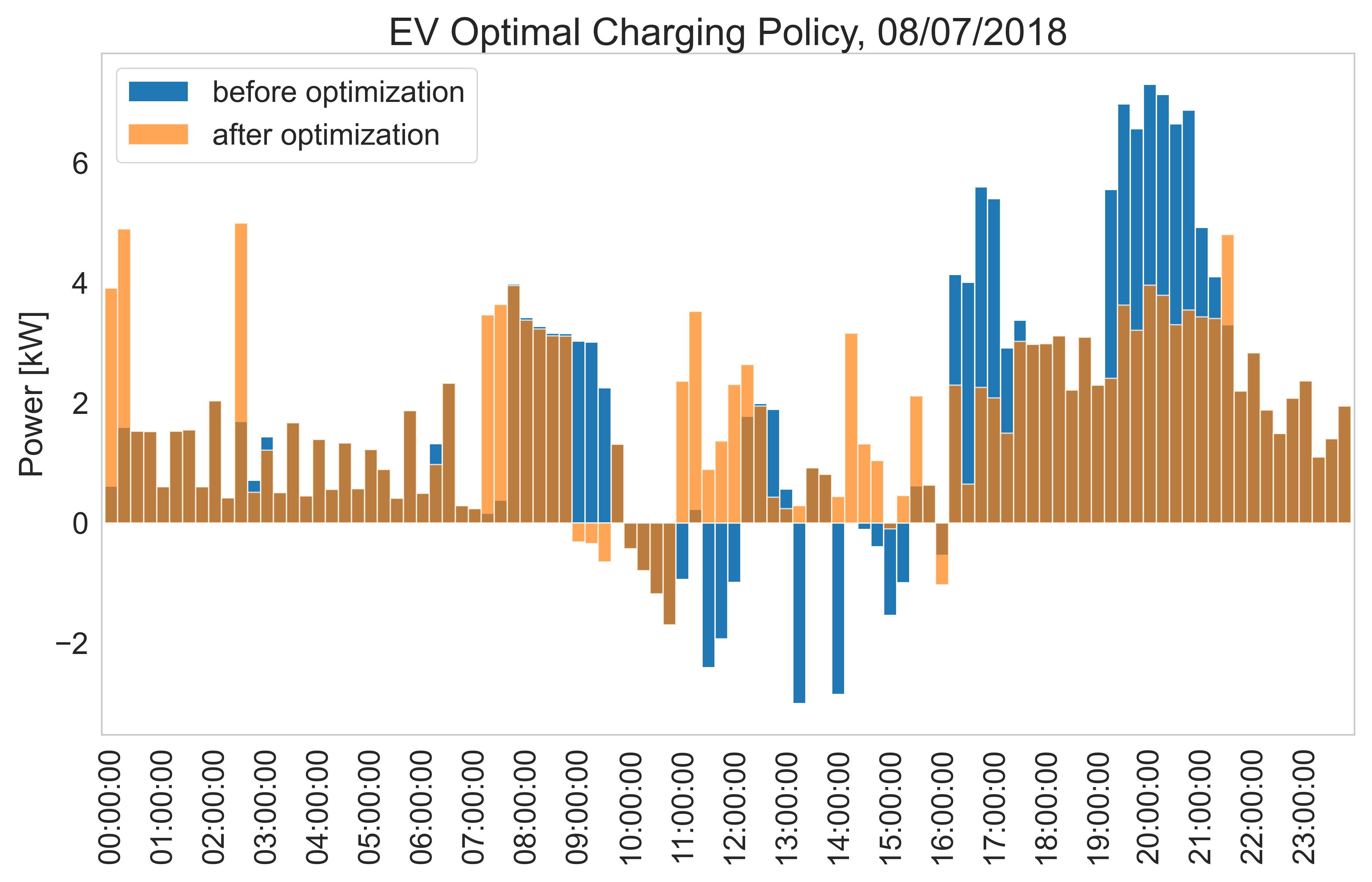}
    \caption{Weekend test case. Comparison between historical charging data and the proposed controlled EV charging policy. The system stress is being reduced due to lower power consumption on On-Peak hours.}
    \label{fig:res_weekend_bar}
\end{figure}

%% file: Main.bbl
\begin{thebibliography}{10}

\bibitem{Karafotis2020Evaluation}
Karafotis, P.A., Evangelopoulos, V.A. and Georgilakis, P.S.: `Evaluation of
  harmonic contribution to unbalance in power systems under non-stationary
  conditions using wavelet packet transform', \emph{Electric Power Systems
  Research},  2020, \textbf{178}, pp.~106026

\bibitem{IEA_EV}
`{Global EV Outlook 2022}'. (International Energy Agency),  2022.
\newblock [Online; accessed 19-July-2022].
\newblock \url{https://www.iea.org/reports/global-ev-outlook-2022}

\bibitem{Sykiotis2022electricity}
Sykiotis, S., Kaselimi, M., Doulamis, A. and Doulamis, N.: `Electricity: An
  efficient transformer for non-intrusive load monitoring', \emph{Sensors},
  2022, \textbf{22}, (8)

\bibitem{OverviewSoC}
Eleftheriadis, P. and Leva, S.: `An overview of data-driven methods for the
  online state of charge estimation'. 2022 IEEE International Conference on
  Environment and Electrical Engineering and 2022 IEEE Industrial and
  Commercial Power Systems Europe (EEEIC / I CPS Europe),  2022.

\bibitem{Rajasekhar2020Survey}
Rajasekhar, B., Tushar, W., Lork, C., Zhou, Y., Yuen, C., Pindoriya, N.M.,
  et~al.: `A survey of computational intelligence techniques for
  air-conditioners energy management', \emph{IEEE Transactions on Emerging
  Topics in Computational Intelligence},  2020, \textbf{4}, (4), pp.~555--570

\bibitem{Mabina2021Sustainability}
Mabina, P., Mukoma, P. and Booysen, M.J.: `Sustainability matchmaking: Linking
  renewable sources to electric water heating through machine learning',
  \emph{Energy and Buildings},  2021, \textbf{246}, pp.~111085

\bibitem{Antonopoulos2020Artificial}
Antonopoulos, I., Robu, V., Couraud, B., Kirli, D., Norbu, S., Kiprakis, A.,
  et~al.: `Artificial intelligence and machine learning approaches to energy
  demand-side response: A systematic review', \emph{Renewable and Sustainable
  Energy Reviews},  2020, \textbf{130}, pp.~109899

\bibitem{Menos2022Particle}
{Menos-Aikateriniadis}, C., Lamprinos, I. and Georgilakis, P.S.: `Particle
  swarm optimization in residential demand-side management: A review on
  scheduling and control algorithms for demand response provision',
  \emph{Energies},  2022, \textbf{15}, (6)

\bibitem{Liu2020Optimization}
Liu, Y., Zhang, D. and Gooi, H.B.: `Optimization strategy based on deep
  reinforcement learning for home energy management', \emph{CSEE Journal of
  Power and Energy Systems},  2020, \textbf{6}, (3), pp.~572--582

\bibitem{Vazquez2019Reinforcement}
Vázquez.Canteli, J.R. and Nagy, Z.: `Reinforcement learning for demand
  response: A review of algorithms and modeling techniques', \emph{Applied
  Energy},  2019, \textbf{235}, pp.~1072--1089

\bibitem{Li2020SafeDQN}
Li, H., Wan, Z. and He, H.: `Constrained ev charging scheduling based on safe
  deep reinforcement learning', \emph{IEEE Transactions on Smart Grid},  2020,
  \textbf{11}, (3), pp.~2427--2439

\bibitem{Wan2019EV_DRL}
Wan, Z., Li, H., He, H. and Prokhorov, D.: `Model-free real-time ev charging
  scheduling based on deep reinforcement learning', \emph{IEEE Transactions on
  Smart Grid},  2019, \textbf{10}, (5), pp.~5246--5257

\bibitem{Li2020DR}
Li, H., Wan, Z. and He, H.: `Real-time residential demand response', \emph{IEEE
  Transactions on Smart Grid},  2020, \textbf{11}, (5), pp.~4144--4154

\bibitem{Chis2017PEV}
Chiş, A., Lundén, J. and Koivunen, V.: `Reinforcement learning-based plug-in
  electric vehicle charging with forecasted price', \emph{IEEE Transactions on
  Vehicular Technology},  2017, \textbf{66}, (5), pp.~3674--3684

\bibitem{Shuvo2022HERS}
Shuvo, S.S. and Yilmaz, Y.: `Home energy recommendation system (hers): A deep
  reinforcement learning method based on residents’ feedback and activity',
  \emph{IEEE Transactions on Smart Grid},  2022, \textbf{13}, (4),
  pp.~2812--2821

\bibitem{Ren2022Novel}
Ren, M., Liu, X., Yang, Z., Zhang, J., Guo, Y. and Jia, Y.: `A novel
  forecasting based scheduling method for household energy management system
  based on deep reinforcement learning', \emph{Sustainable Cities and Society},
   2022, \textbf{76}, pp.~103207

\bibitem{Pecan}
`{Dataport - Pecan Street Inc.}'. (Dataport),  2009.
\newblock [Online; accessed 19-July-2022].
\newblock \url{https://www.pecanstreet.org/dataport/}

\bibitem{synergy22}
Sykiotis, S., Menos.Aikateriniadis, C., Doulamis, A., Doulamis, N. and
  Georgilakis, P.S.: `Solar power driven ev charging optimization with deep
  reinforcement learning'. 2022 2nd International Conference on Energy
  Transition in the Mediterranean Area (SyNERGY MED),  2022. pp.~ 1--6

\bibitem{AustinPrices}
`{City of Austin Fiscal Year 2018 Electric Tariff}'. ({City of Austin}),  2018.
\newblock [Online; accessed 19-July-2022].
\newblock \url{https://www.austintexas.gov/edims/document.cfm?id=282288}

\bibitem{NissanLeaf}
`{Steady State Vehicle Charging Fact Sheet: 2015 Nissan Leaf}'. ({Idaho
  National Laboratory}),  2015.
\newblock [Online; accessed 19-July-2022].
\newblock
  \url{https://avt.inl.gov/sites/default/files/pdf/fsev/SteadyStateLoadCharacterization2015Leaf.pdf}

\end{thebibliography}
